\pdfoutput=1

\documentclass[11pt]{article}

\usepackage{ACL2023}

\usepackage{times}
\usepackage{latexsym}

\usepackage[T1]{fontenc}

\usepackage[utf8]{inputenc}

\usepackage{inconsolata}

\usepackage{hyperref}       
\usepackage{bbm}
\usepackage{url}            
\usepackage{booktabs}       
\usepackage{amsfonts}       
\usepackage{nicefrac}       
\usepackage{microtype}      
\usepackage{xcolor}         
\usepackage{enumitem}
\usepackage{amsmath}
\usepackage{booktabs}
\usepackage{thm-restate}
\usepackage{amssymb}
\usepackage{graphicx}
\usepackage{subcaption}

\title{C3: Zero-shot Text-to-SQL with ChatGPT}

\author{Xuemei Dong$^{1}$, Chao Zhang$^{1}$, Yuhang Ge$^{1}$, Yuren Mao$^{1}$, Yunjun Gao$^{1}$,\\ \textbf{ Lu Chen$^{1}$, Jinshu Lin$^{2}$, Dongfang Lou$^{2}$} \\
  $^{1}$Zhejiang University, Zhejiang, China \\
  $^{2}$Hundsun Technologies INC., Zhejiang, China \\
  \texttt{\{xm.dong,zjuzhangchao,yuhangge,yuren.mao,gaoyj,luchen\}@zju.edu.cn} \\\texttt{loudongfang2022@gmail.com,linjs13607@hundsun.com} \\
}

\begin{document}
\maketitle
\begin{abstract}
  This paper proposes a ChatGPT-based zero-shot Text-to-SQL method, dubbed C3, which achieves 82.3\% in terms of execution accuracy on the holdout test set of Spider and becomes the state-of-the-art zero-shot Text-to-SQL method on the Spider Challenge. C3 consists of three key components: \underline{C}lear Prompting (CP), \underline{C}alibration with Hints (CH), and \underline{C}onsistent Output (CO), which are corresponding to the model input, model bias and model output respectively. It provides a systematic treatment for zero-shot Text-to-SQL. Extensive experiments have been conducted to verify the effectiveness and efficiency of our proposed method. Our code is available at \href{https://github.com/bigbigwatermalon/C3SQL}{https://github.com/bigbigwatermalon/C3SQL}
\end{abstract}

\section{Introduction}
Text-to-SQL, which aims to convert natural language questions into executable SQL queries, can be used to provide a user-friendly interface to relational databases. It can benefit various aspects of data management, such as accessibility to databases, flexibility of website design and so on. 

Traditional Text-to-SQL methods~\cite{li2023decoupling} typically fine-tune a decoder-encoder model with an amount of training data to achieve proper Text-to-SQL performance. These fine-tuning-based methods require a training set that consists of amounts of text-SQL pairs. However, in practice, obtaining the text-SQL pairs is extremely expensive. Furthermore, this fine-tuning paradigm can cause overfitting of the training set and degenerate the transferability of the model. 

To address these problems, few-shot or zero-shot Text-to-SQL methods are natural choices. However, the decoder-encoder model-based methods cannot achieve proper performance~\cite{Zihui-few-shot}. Recently, the emergent abilities of Large Language Models (LLMs) make few-shot \cite{pourreza2023din} or zero-shot~\cite{liu2023comprehensive} Text-to-SQL possible. The few-shot  Text-to-SQL method~\cite{pourreza2023din} can outperform the fine-tuning-based methods by means of few-shot in-context learning; however, it requires more than 10,000 tokens for each query, which is expensive and infeasible in practical Text-to-SQL tasks. By contrast, the zero-shot Text-to-SQL method~\cite{liu2023comprehensive}  can save tokens; however, its performance is inferior to the fine-tuning-based methods.



To address these issues, this paper proposes a novel ChatGPT-based zero-shot Text-to-SQL method, dubbed C3, which utilizes only approximately 1,000 tokens per query and achieves a better performance than fine-tuning-based methods. C3 consists of three key components: \underline{C}lear Prompting (CP), \underline{C}alibration with Hints (CH) and \underline{C}onsistent Output (CO), which are corresponding to the model input, model bias and model output respectively.  Specifically, CP is a novel prompting paradigm for zero-shot Text-to-SQL, which improves the zero-shot Text-to-SQL performance via adopting proper input. Furthermore, CH is proposed to mitigate the biases of ChatGPT in generating SQL queries, which improves the zero-shot Text-to-SQL performance via calibrating the model biases. Besides, CO is designed to keep the consistency of the generated SQL queries, which improves the zero-shot Text-to-SQL performance via overcoming the inherent randomness and uncertainty in the output of large language models.




We evaluate our proposed C3 method on the widely used Spider dataset, where  C3 achieves 82.3\% in terms of execution accuracy on the holdout test set of Spider and becomes the state-of-the-art zero-shot Text-to-SQL method. It outperforms state-of-the-art fine-tuning-based approaches by 2.4\% execution accuracy on the holdout test set; besides, it only uses approximately 1,000 tokens per query. Furthermore, our extensive experimental analysis provides several noteworthy discoveries, which are potential to forge a new trend in GPT-based Text-to-SQL research.

 

\section{Related Work}
\label{gen_inst}
Existing Text-to-SQL methods can be divided into several categories: \textit{rule-based methods}, \textit{fine-tuning methods} and \textit{ICL-based methods}. The \textit{rule-based Text-to-SQL methods} \cite{DBLP:conf/aaai/ZelleM96,DBLP:journals/pvldb/SahaFSMMO16} use well designed templates to generate SQL queries, which have achieved good performance in specific cases. However, these methods rely heavily on manual rule design, making them difficult to apply to other domains and limiting their scalability and generalizability.

To address these limitations, researchers explore Seq2Seq models based on bidirectional LSTMs \cite{DBLP:conf/acl/GuoZGXLLZ19} and CNNs \cite{DBLP:journals/coling/ChoiSKS21}. These models offer improved flexibility and effectiveness but face challenges in integrating database structural information seamlessly. To tackle these constraints, researchers utilize graph neural networks, treating the database schema as a graph, to better integrate structural information \cite{DBLP:conf/acl/WangSLPR20, DBLP:conf/acl/CaoC0ZZ020}. Besides, fine-tuning on pretrained language models like T5 has demonstrated better performance \cite{DBLP:journals/jmlr/RaffelSRLNMZLL20, DBLP:conf/emnlp/ScholakSB21, li2023decoupling}. However, fine-tuning methods typically require a substantial volume of labeled training data specific to the target task and they often suffer from overfitting to the training data.

The emergence of large language models (LLM) has left a deep impression and points towards new directions for the Text-to-SQL task. LLMs like GPT models outperform fine-tuning models on many NLP down-stream tasks in few-shot or zero-shot setting due to the ability of in-context learning (ICL) \cite{brown2020language}. The design of prompts in the ICL profoundly influences the quality of outputs \cite{DBLP:conf/emnlp/MinLHALHZ22, DBLP:conf/acl-deelio/LiuSZDCC22, DBLP:conf/nips/Wei0SBIXCLZ22} in LLM. Some work \cite{rajkumar2022evaluating, liu2023comprehensive} conduct the evaluation of LLMs' ICL performance on Text-to-SQL task with different prompts. However, none of them outperform the current fine-tuning methods. 

All though the standard prompting only provides a lower bound on the capabilities of LLMs in principle \cite{DBLP:conf/nips/Wei0SBIXCLZ22}, LLMs with well designed prompts have the potential to achieve better performance. Recently, a few-shot paradigm using GPT-4 has achieved SOTA performance on Text-to-SQL task \cite{pourreza2023din}. But this method needs some handcraft demonstrations and numerous tokens, which are time-consuming and resource-intensive. 

In our work, we demonstrate that ChatGPT can effectively generate SQL queries without the need for any demonstrations when provided with proper instructions. We show its capability as a skilled zero-shot SQL writer. 

\begin{figure*}
  \centering
  \includegraphics[width=1.0\linewidth]{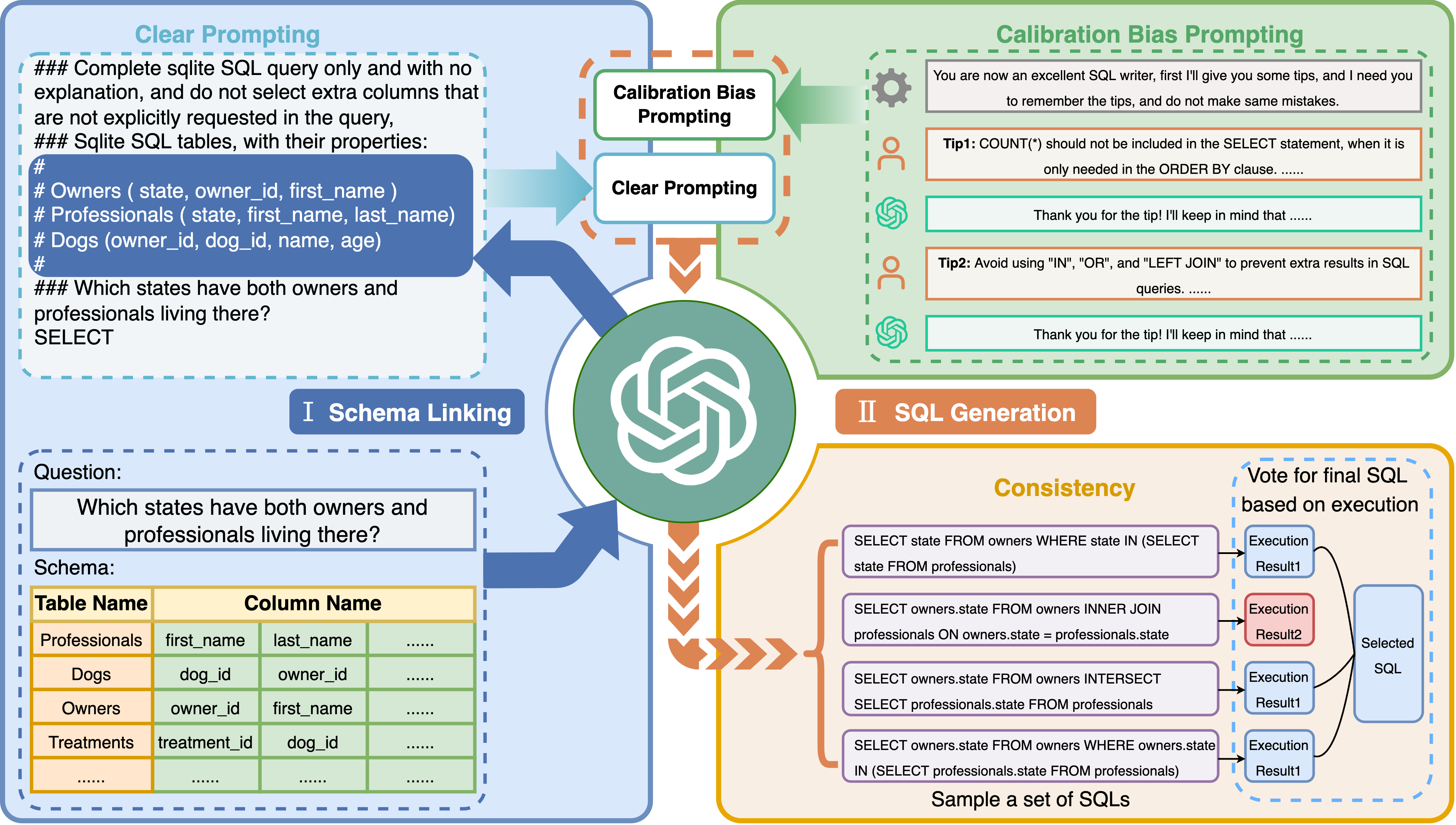}
  \caption{The framework of C3.}
  \label{fig:framework}
\end{figure*}

\section{Preliminaries}
\label{preliminaries}
\subsection{Problem Definition of Text-to-SQL}
Given natural language question $Q$ and database schema $\mathcal{S}$, the database schema $\mathcal{S}=\{\mathcal{T}, \mathcal{C}, \mathcal{R}\}$ includes multiple tables $\mathcal{T}$, columns $\mathcal{C}$ and foreign key relations $\mathcal{R}$. The problem of Text-to-SQL parsing aims to generate the SQL query $\mathcal{Y}$ corresponding to the question.

\subsection{Large Language Model for Text-to-SQL}
The task of Text-to-SQL parsing has been formulated as a generation task in recent works~\cite{sun2023sql,liu2023comprehensive}, utilizing appropriate prompts $\mathcal{P}$ to guide a large language model $\mathcal{M}$. This model estimates a probability distribution over SQL queries $\mathcal{Y}$ and allows us to generate queries token by token. The generation process for the SQL query $\mathcal{Y}$ can be formulated as follows:
\begin{equation}
P_{\mathcal{M}}(\mathcal{Y} | \mathcal{P}, \mathcal{S}, Q) = \prod_{i=1}^{|\mathcal{Y}|} P_{\mathcal{M}}(\mathcal{Y}_i | \mathcal{P}, \mathcal{S}, Q, \mathcal{Y}{<i})
\end{equation}

Here, $\mathcal{Y}{<i}$ is the prefix of the SQL query $\mathcal{Y}$ and $P_{\mathcal{M}}(\mathcal{Y}_i | \cdot)$ is the conditional probability of the $i$-th token in the SQL query $\mathcal{Y}$ given the prefix $\mathcal{Y}_{<i}$, the prompt $\mathcal{P}$, the schema $\mathcal{S}$ and question $\mathcal{Q}$.


Recent studies have demonstrated that large language models (LLMs) can learn from a few examples within a given context, known as in-context learning~\cite{sun2023sql,DBLP:conf/emnlp/MinLHALHZ22,pourreza2023din}. These works have shown that in-context learning effectively enables LLMs to perform a range of complex tasks. However, including additional examples tends to increase the manual cost and token cost associated with using the OpenAI API. Therefore, in this study, we specifically focus on the zero-shot prompt setting.

\section{Proposed Approach}
\label{method}
In this section, we propose the C3 method which can achieve proper zero-shot Text-to-SQL performance based on ChatGPT. The framework of the C3 method is demonstrated in Figure~\ref{fig:framework}. C3 consists of three key components: \underline{C}lear Prompting (CP), \underline{C}alibration with Hints (CH), and \underline{C}onsistency Output (CO), which are corresponding to the model input, model bias and model output respectively. The details of each component are introduced as follows.

\subsection{Clear Prompting}
\label{cp}


The Clear Prompting (CP) component aims to provide effective prompts for Text-to-SQL parsing. It consists of two parts: clear layout and clear context.


\paragraph{Clear Layout}




In Text-to-SQL, the widely used prompt layout styles can be divided into two types. We denote them as complicated layout and clear layout respectively. 
The details of each type are introduced as follows.  

\begin{figure}[ht]
  \begin{minipage}[t]{1.0\linewidth}
    \centering
    \includegraphics[width=1.0\textwidth]{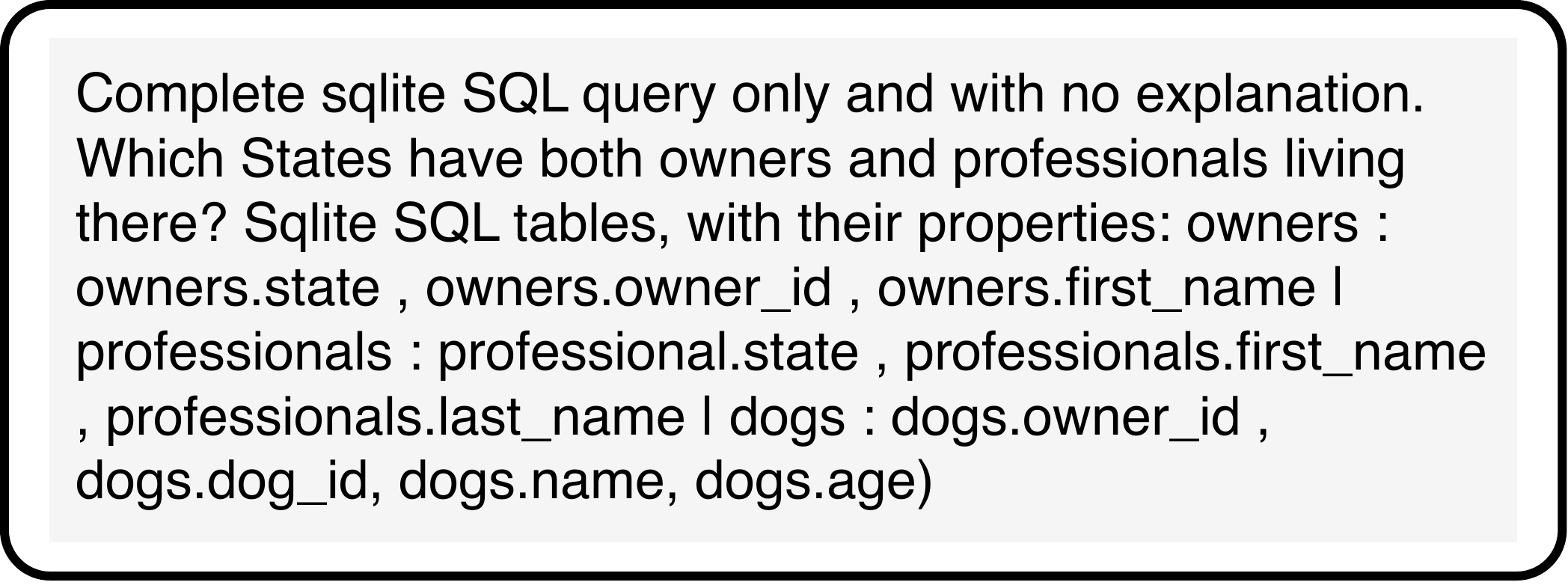}
    \subcaption{Traditional PLM prompt.}
    \label{fig:comparing_a}\hspace{0.1cm}
  \end{minipage}
  
  \begin{minipage}[t]{1.0\linewidth}
    \centering
    \includegraphics[width=1.0\textwidth]{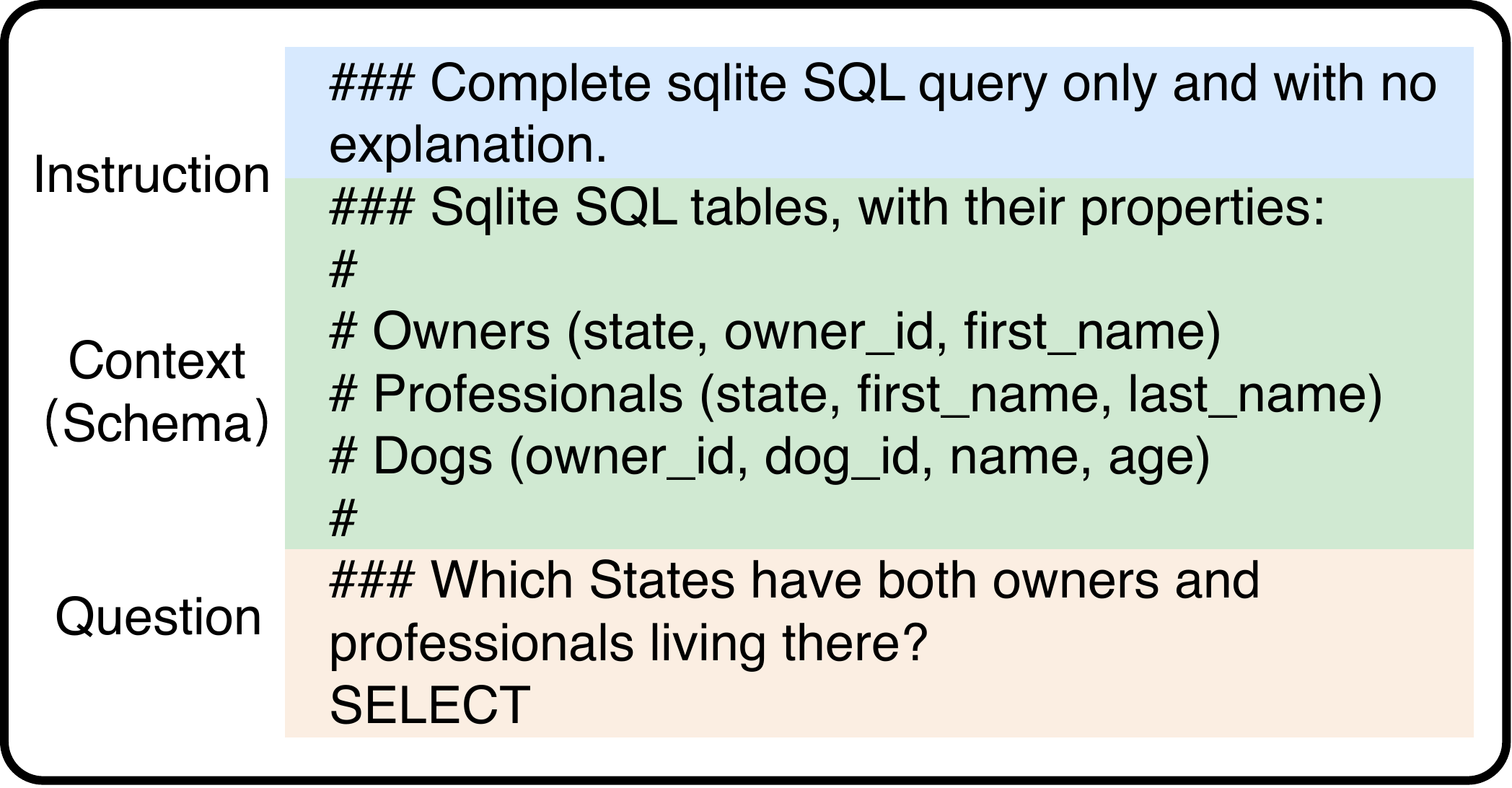}
    \subcaption{OpenAI prompt.}
    \label{fig:comparing_b}\hspace{0.1cm}
  \end{minipage}
  
  \begin{minipage}[t]{1.0\linewidth}
    \centering
    \includegraphics[width=1.0\textwidth]{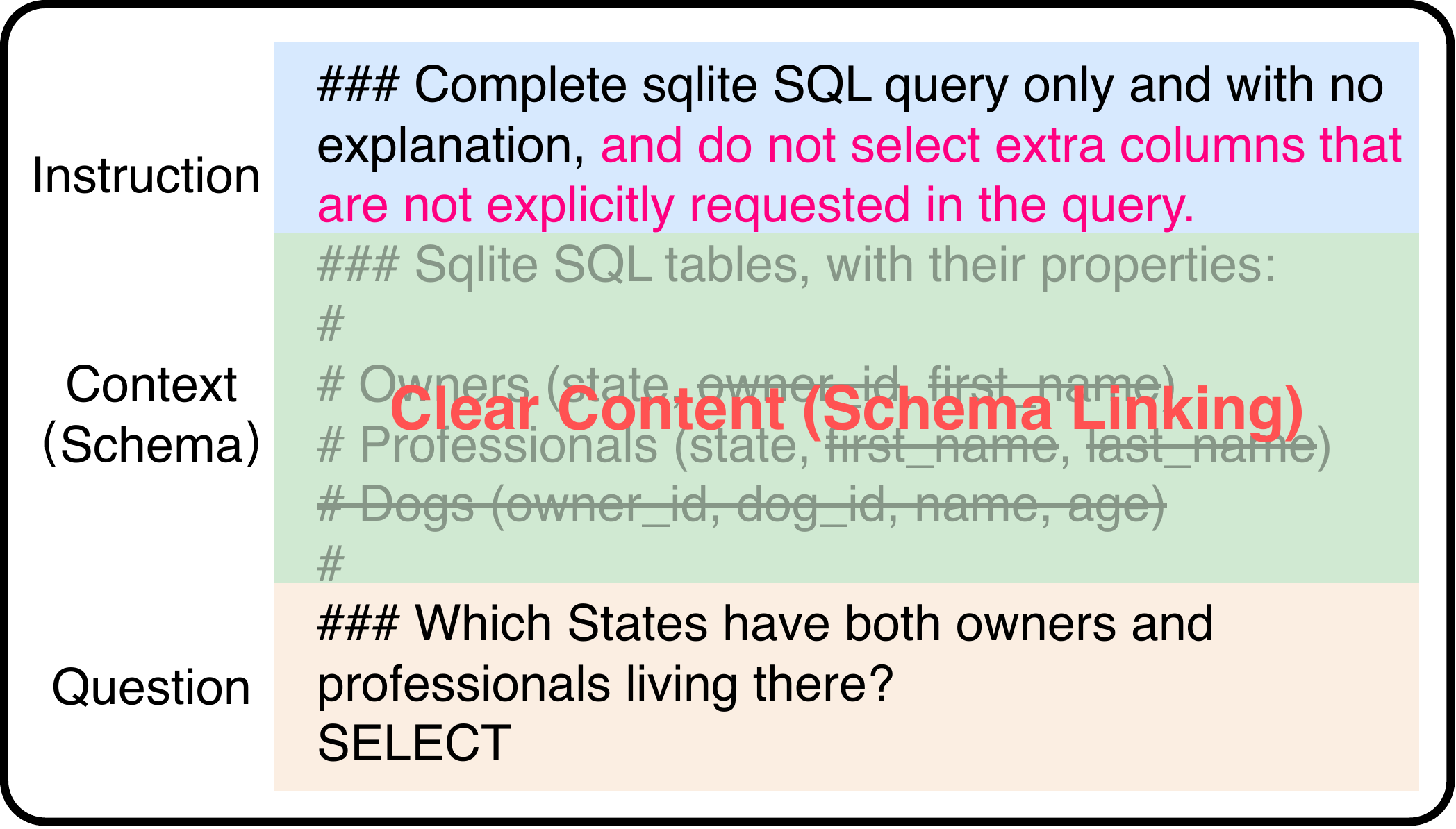}
    \subcaption{Our proposed C3 prompt.}
    \label{fig:comparing_c}
  \end{minipage}
  \caption{The comparison of Traditional PLM prompt, OpenAI prompt and C3 prompt.}  
\end{figure}

\begin{itemize}
  \item \textbf{Type 1: Complicated Layout}: This type of prompt layout directly concatenates the instruction, question and context (database schema) together, which looks a mess. An example of this prompt layout is illustrated in Figure~\ref{fig:comparing_a}.
  
  \item \textbf{Type 2: Clear Layout}: This type of layout divides the instruction, context (database schema) and questions by adopting sharp symbols, which look clear. An example of this prompt layout is illustrated in Figure~\ref{fig:comparing_b}.
\end{itemize}

Intuitively, the clear layout will be easily understood by ChatGPT and achieve better performance. The experiments illustrate that the clear layout outperforms the complicated layout by 7.0\% in term of execution accuracy, which verify this intuition.  
The experimental setting can be found in Section \ref{sec:effect_of_clear_prompt}). Therefore, this paper adopts a clear layout to construct our prompt template.
We also provide an interesting material that asks ChatGPT itself whether clear layout is important (please refer to Appendix \ref{sec:conversation}).  However, our experimental results show that directly using the instruction provided in Figure~\ref{fig:comparing_b} causes redundant columns in the generated SQL.
To solve this problem, we improve the instruction part by adding a clause \textit{and do not select extra columns that are not explicitly requested in the query}.

Moreover, the context provided in Figure~\ref{fig:comparing_b}  requires to refine. We refine it in the following section.

\paragraph{Clear Context}



Including the entire database schema in the context part of a prompt causes two issues. Firstly, introducing too many irrelevant items in the prompt increases the likelihood of ChatGPT generating irrelevant schema items in the output SQL. Secondly, using the complete database schema results in excessive text length, leading to unnecessary API costs. To overcome these two issues, we propose that it is necessary to conduct schema linking which recalls relevant tables and columns. With the schema linking results, we can only put the linked tables and columns into the context part. This paper presents a ChatGPT-based zero-shot schema linking approach, which can be divided into the following two steps:

\begin{itemize}
    \item \textbf{Table Recall} We design a zero-shot prompt to instruct ChatGPT to recall tables using three steps. Firstly, the tables should be ranked based on their relevance to the question. Secondly, the model should check if all tables have been considered. Finally, the output format is specified as a list. To ensure the stability of table recall, we employ a self-consistency method. Specifically, the model generates ten sets of retrieval results, each set containing the top four tables. The final result is determined by selecting the set that appears most frequently among the ten sets. The complete prompt is shown in Appendix ~\ref{sec:table_recall}.
    \item \textbf{Column Recall} Based on the table recall results, we further retrieve the columns within the candidate tables. We also design a zero-shot prompt and instructed ChatGPT to recall columns in two steps. Firstly, all columns within each candidate table are ranked based on their relevance to the question. Then, the output format is specified as a dictionary. In the prompt, we also emphasize that columns matching more with the question words or the foreign key should be placed ahead to assist in more accurate recall results. Similarly, we employ the self-consistency method. Specifically, for each table, we generate ten sets of recalled columns at once. Then we choose five columns that appear most frequently among each set as the final result. The complete prompt is shown in Appendix ~\ref{sec:column_recall}.
\end{itemize}

Besides the recalled tables and columns, we also add foreign key information of the recalled tables into the context part to specify the columns required for \textit{JOIN} operations. 

Combining the Clear Layout and Clear Context, we propose Clear Prompt, as demonstrated in Figure \ref{fig:comparing_c}.

\subsection{Calibration of Model Bias}
\label{CH}
\begin{figure*}
  \centering
  \includegraphics[width=1.0\linewidth]{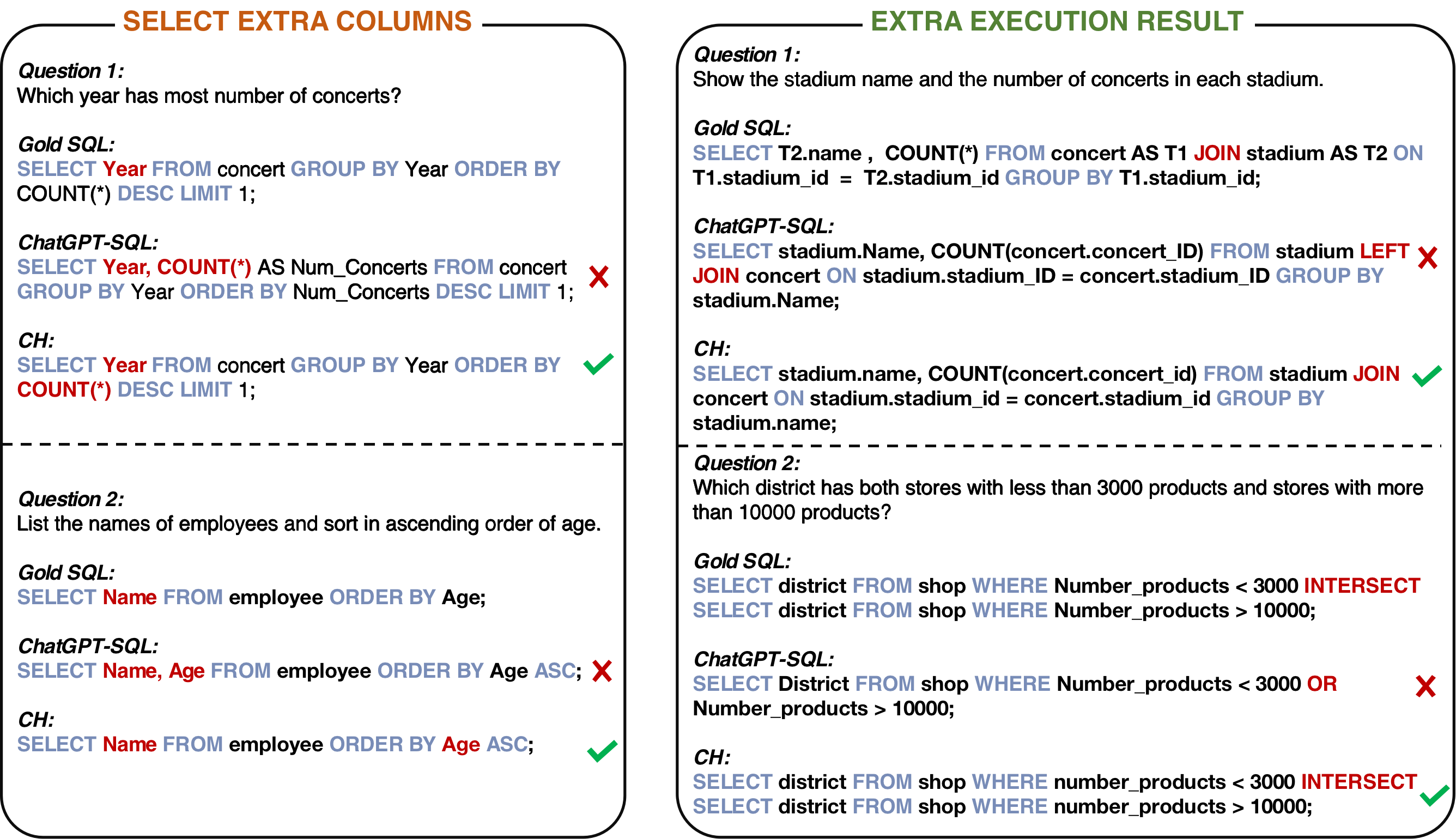}
  \caption{The examples of mistakes caused by the model's biases.}
  \label{fig:error_example}
\end{figure*}

Through analyzing the errors that occurred in the generated SQL queries, we find that some errors are caused by certain biases inherent in ChatGPT. As demonstrated in Figure~\ref{fig:error_example}, ChatGPT prefers to provide extra columns and extra execution results. This paper concludes them as the following two kinds of bias.

\begin{itemize}
    \item \textbf{Bias 1}: ChatGPT tends to be conservative in its output and often selects columns that are relevant to the question but not necessarily required. Furthermore, we find that this tendency is particularly pronounced when it comes to issues involving quantities. For example, for the first question in Figure~\ref{fig:error_example} (left), ChatGPT chooses \textit{Year} and \textit{COUNT(*)} in the \textit{SELECT} clause. However, the gold SQL in Spider dataset only selects \textit{Year} as \textit{COUNT(*)} is only needed for ordering purposes.
    \item \textbf{Bias} 2: ChatGPT tends to use \textit{LEFT JOIN}, \textit{OR} and \textit{IN} when writing SQL queries, but often fails to use them correctly. This bias often leads to extra values in execution results. Some examples of this bias can be found in Figure~\ref{fig:error_example} (right).
\end{itemize}

To calibrate these two biases, we propose a plug-in calibration strategy, dubbed \textit{Calibration with Hints (CH)}. CH incorporates prior knowledge into ChatGPT by using contextual prompts which include historical conversations. 
In the historical conversation, we initially regard ChatGPT as an excellent SQL writer and guide it to follow our proposed debias hints. 
\begin{itemize}
    \item \textbf{Hint 1}: For the first bias, we design a tip to guide ChatGPT in selecting only the necessary column. This tip is illustrated in the upper right part of Figure~\ref{fig:framework}. It emphasizes that items like \textit{COUNT(*)} should not be included in \textit{SELECT} clause when it is only needed for ordering purposes. 
    \item \textbf{Hint 2}: For the second bias, we design a tip to prevent ChatGPT from misusing SQL keywords. As shown in the upper right part of Figure~\ref{fig:framework}, we straightforwardly ask ChatGPT to avoid using \textit{LEFT JOIN}, \textit{IN} and \textit{OR}, and use \textit{JOIN} and \textit{INTERSECT} instead. We also instruct ChatGPT to use \textit{DISTINCT} or \textit{LIMIT} when appropriate to avoid repetitive execution results.
\end{itemize}
 By incorporating these two hints, ChatGPT can generate SQL queries that align more closely with the desired output. As shown in Figure~\ref{fig:error_example}, using our CH prompt can effectively calibrate the model biases. More examples of CH prompt can be found in Appendix ~\ref{sec:c3_prompt}.

\subsection{Consistency Output}
\label{CO}
\begin{figure*}
  \centering
  \includegraphics[width=\textwidth]{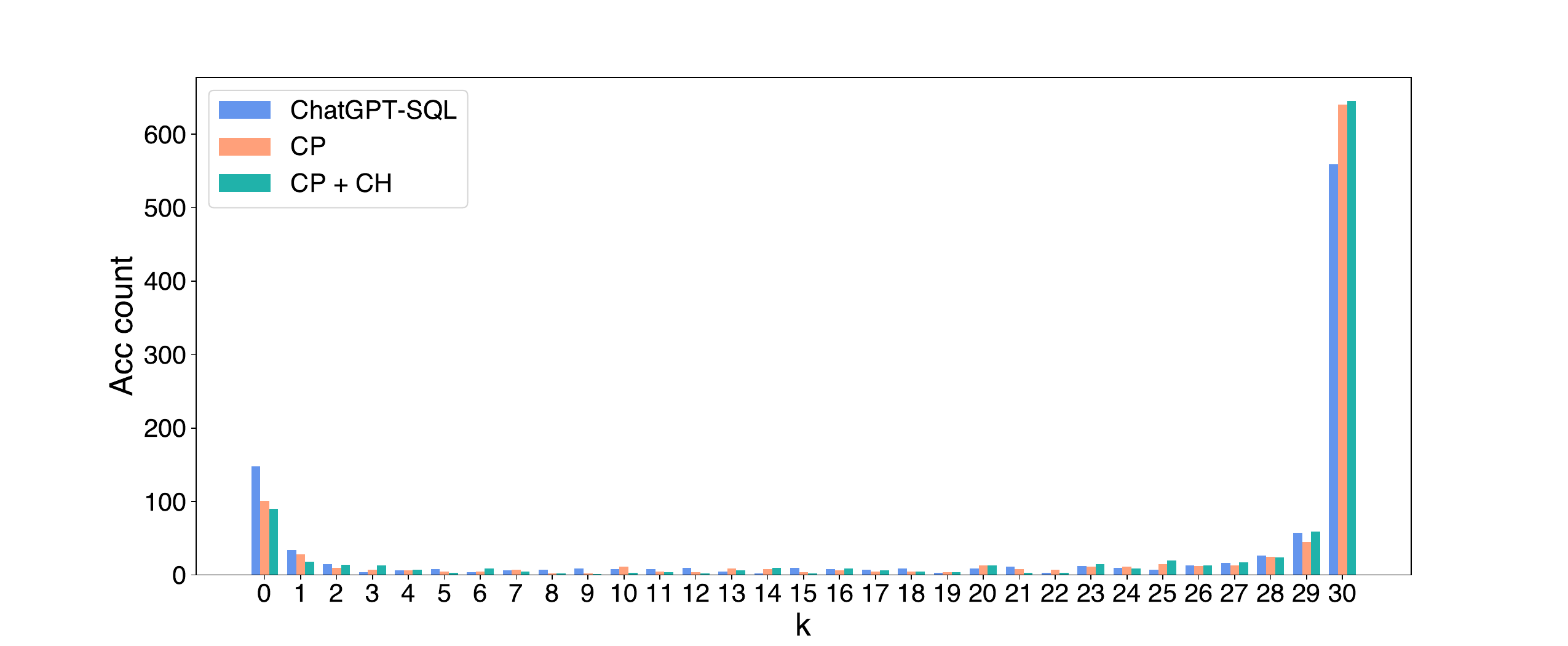}
  \caption{The distribution of correctness counts on the dev set across thirty independent experiments under different prompts. The horizontal axis represents the number of correctness, and the vertical axis represents the number of samples achieving k correct results out of 30 trials.}
  \label{fig:acc_count}
\end{figure*}

\begin{figure}[t]
  \centering
  \includegraphics[width=0.95\linewidth]{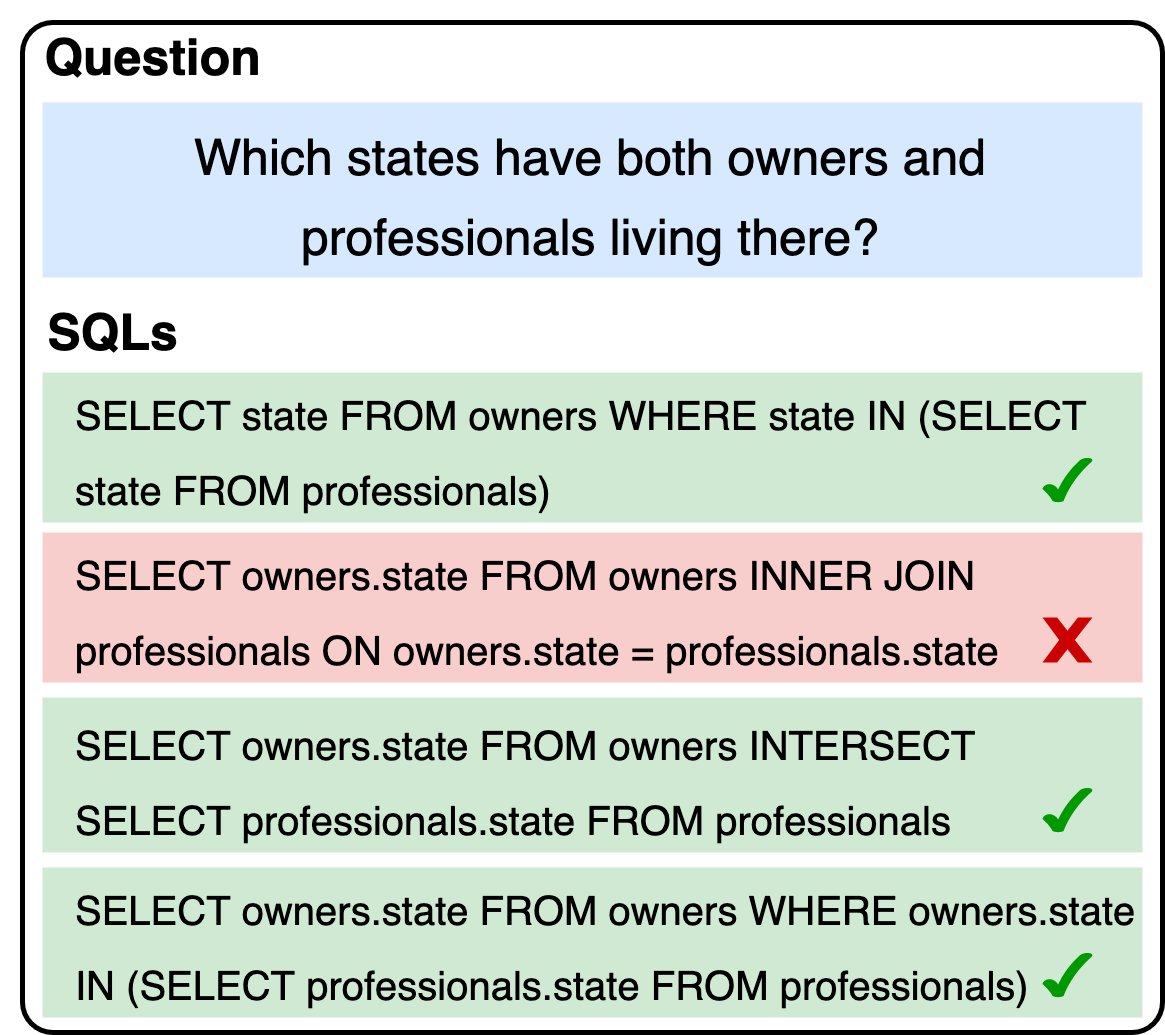}
  \caption{A example of multiple different SQL queries for the same question.}
  \label{fig:sc_ex}
\end{figure}

Using the CP and CH methods, ChatGPT is able to generate higher-quality SQL. However, the output of ChatGPT is unstable due to the inherent randomness of large language models~\cite{lin2022teaching}. To find out the influence of the uncertainty output of ChatGPT, we analyze the distribution of correctness counts on the dev set across thirty independent experiments under different prompts, as shown in Figure~\ref{fig:acc_count}. In this figure,  ChatGPT-SQL is the method proposed in literature~\cite{liu2023comprehensive}; besides, CP and CP + CH denote our proposed Clear Prompt and the combination of Clear Prompt and Clear Hint method respectively.
Regardless of the method used, only less than 65\% of SQL statements can consistently be written correctly.
This means that by enhancing the consistency of the output, the model has a great potential to correctly write the majority of the queries.

To enhance the consistency, we draw inspiration from the concept of \textit{self-consistency}~\cite{wang2023selfconsistency} proposed in previous work. \textit{Self-consistency} method is motivated by the fact that in complex reasoning problems, there are multiple different reasoning paths to the unique right answer. It first samples multiple different reasoning paths and then selects the most consistent answer to improve the quality of the output remarkably. Text-to-SQL problem is similar to reasoning problems, where there are multiple ways to write SQL queries to represent the same meaning as shown in the figure~\ref{fig:sc_ex}. Therefore, we implement execution-based \textit{self-consistency} to Text-to-SQL. 

Specifically, we first sample multiple reasoning paths to generate diverse SQL answers. Then, execute these SQL queries on the database and collect the execution outcomes. After removing the errors from all the outcomes, we identify the most consistent SQL as the final SQL by applying a voting mechanism to these execution results. For instance, in figure~\ref{fig:sc_ex}, we classify the SQL queries based on the execution outcomes and represent them with different colors. Then we compare the categories to determine which one contains more SQL queries, and select one SQL from that category as the final SQL. This approach allows us to harness the collective knowledge derived from these multiple paths, resulting in more reliable and resilient results in the generation of SQL queries.

\section{Experiments}

\begin{table*}[!]
  \caption{Comparison of previous methods and our method in terms of execution accuracy on Spider dataset.}
  \label{tab: overall results}
  \centering
  \begin{tabular}{c|ccc|cc}
    \toprule
    Model & Zero-Shot & Few-Shot & Fine-tuning & Dev & Test \\
    \midrule
    ChatGPT-SQL + ChatGPT \cite{liu2023comprehensive} & \checkmark &  &  & 72.3 & - \\
    RATSQL + GAP + NatSQL \cite{DBLP:conf/acl/WangSLPR20} &  &  & \checkmark & 75.0 & 73.3 \\
    T5-3B + PICARD \cite{DBLP:conf/emnlp/ScholakSB21} &  &  & \checkmark & 79.3 & 75.1 \\
    Graphix-3B + PICARD \cite{li2023graphix} &  &  & \checkmark & 81.0 & 77.6 \\
    SC-Prompt + T5-3B \cite{Zihui-few-shot} &  & \checkmark & \checkmark & 81.1 & - \\
    RESDSQL-3B + NatSQL \cite{li2023decoupling} &  &  & \checkmark & 84.1 & 79.9 \\
    DIN-SQL + GPT-4 \cite{pourreza2023din} &  & \checkmark &  & 82.8 & 85.3 \\
    \hline
    C3 + ChatGPT (Ours) & \checkmark &  &  & 81.8 & 82.3 \\ 
    \bottomrule
  \end{tabular}
\end{table*}

\subsection{Experimental Setup}
\textbf{Datasets} 
We conduct experiments on the Spider dataset. Spider \cite{DBLP:conf/emnlp/YuZYYWLMLYRZR18} is a large-scale complex and cross-domain Text-to-SQL dataset. It consists of 10,181 questions and 5,693 unique complex SQL queries on 200 databases with multiple tables covering 138 different domains. It is divided into 8659 training samples across 146 databases, 1034 evaluation samples across 20 databases and 2147 test samples across 34 databases. There is no overlap between the databases used in each of these sets. Meanwhile, according to their difficulty levels, these samples are classified into four categories: easy, medium, hard, and extra.

\textbf{Evaluation Metrics}
The most commonly used evaluation metrics are exact match (EM) and execution accuracy (EX). For EM, it considers each part of the SQL query predicted by the model after removing the values, and it is considered a correct prediction only if all parts match the corresponding parts in the standard SQL query. For EX, it evaluates the correctness of the predicted SQL query by comparing the execution results of the predicted SQL and the standard SQL in the same database. In our experiment, we only use EX as the evaluation metrics\cite{ruiqi20}, because we think that the SQL can be written in multiple ways to generate the same result. Moreover, ChatGPT has its own style of writing SQL queries. Thus it is more meaningful to make a direct comparison of the results in our experiment. 

\textbf{Implementation}
We use OpenAI ChatGPT API: GPT-3.5-Turbo-0301 as our experiment model. We recall tables and columns in two steps with ChatGPT on their relevance according to the given question. Then we construct input prompt using our CP and CH methods. Lastly, we utilize ChatGPT to generate 20 SQL queries at once and select the most consistent SQL as the final result using our CO method.  

\textbf{Baselines}
We conduct experiments on the Spider dataset and compare our method with the following baselines:
\begin{itemize}
    \item \textbf{ChatGPT-SQL} \cite{liu2023comprehensive} introduces a simple zero-shot method to evaluate the capability of ChatGPT on the Text2SQL task.
    \item \textbf{RATSQL} \cite{DBLP:conf/acl/WangSLPR20} proposes a unified framework utilizing the relation-aware self-attention to encode the relational structure of a database schema and the query.
    \item \textbf{PICARD} \cite{DBLP:conf/emnlp/ScholakSB21} introduces an incremental parsing method with constrained decoding, which checks and filters out inadmissible tokens at each generation step.
    \item \textbf{Graphix} \cite{li2023graphix} designs a Graphix layer to encode a combination of semantic and structural information based on the pre-trained T5 model.
    \item \textbf{SC-Promp} \cite{Zihui-few-shot} decomposes the Text2SQL task into two subtasks, predicting structure and content separately, achieving high performance with fewer training samples.
    \item \textbf{RESDSQL} \cite{li2023decoupling} proposes a ranking-enhanced encoding and skeleton-aware decoding framework to decouple the schema linking and the skeleton parsing. It is the best approach based on fine-tuning.
    \item \textbf{DIN-SQL} \cite{pourreza2023din} decomposes the text-to-SQL task into smaller subtasks and designs different prompts for each subtask to instruct GPT-4 to complete each subtask and obtain the final SQL. It is the SOTA at the time of writing.
\end{itemize}

\subsection{Overall Performance}

In Table~\ref{tab: overall results}, we report the performance of our method and baseline methods on the Spider dataset.  It can be observed that our method outperforms all traditional fine-tuning-based methods in terms of execution accuracy on the test set. Our method also outperforms the zero-shot setting of ChatGPT-SQL by 9.5\% on the dev set. At the time of writing, our model achieved 2rd on the Spider Leaderboard. Compared with the few-shot setting of top-1 method, DIN-SQL, our method focuses on the zero-shot setting. Our approach only uses approximately 10\% of the token numbers of DIN-SQL; furthermore, the cost of GPT-3.5 engine is also significantly lower than GPT-4 which is used in DIN-SQL. Therefore, our method is more budget-friendly.

\subsection{Effect of Clear Prompt}
\label{sec:effect_of_clear_prompt}
In this section, we experimentally investigate the impact of the prompt layout and the content on the performance of Text-to-SQL respectively. The results are the average execution accuracy obtained on the dev set over five runs.

\begin{table}[t]
  \caption{Comparison with different prompt layout}
  \label{tab: different prompt}
  \centering
  \begin{tabular}{ccc}
    \toprule
    Technique & Dev Acc \\
    \midrule
    Clear Layout & 72.3 \\
    Complicated Layout & 65.3 (↓ 7.0) \\
    \bottomrule
  \end{tabular}
\end{table}

\begin{table}[t]
  \caption{Ablation Study}
  \label{tab: ablation}
  \centering
  \begin{tabular}{cc}
    \toprule
    Technique & Dev Acc \\
    \hline
    C3 + ChatGPT & 81.8 \\
    \hline
    w/o Recall of Tables and Columns & 79.5 \\
    w/o Foreign Keys & 79.2 \\
    w/o Calibration & 80.3 \\
    w/o Self-Consistency & 80.5 \\
    \bottomrule
  \end{tabular}
\end{table}

\begin{table}[t]
  \caption{AUC of Table and Column Recall}
  \label{tab: AUC}
  \centering
  \begin{tabular}{ccc}
    \toprule
    Table AUC & Column AUC \\
    \midrule
    0.9588 & 0.9833 \\
    \bottomrule
  \end{tabular}
\end{table}

\textbf{Clear Layout} We first explore the effect of clear prompt layout by comparing the performance of the clear layout (refer to Type 2 in Section \ref{cp}) and the complicated layout (refer to Type 1 in Section \ref{cp}). The results are recorded in  Table~\ref{tab: different prompt}. From this table, we can see that  clear layout outperforms complicated layout by 7.0\%. It demonstrates that the clear prompt layout is important for chatGPT-based zero-shot Text-to-SQL. We suppose that this is because a clear layout allows ChatGPT to better understand the prompt, while the complicated one may blur the information. So a clear layout can help ChatGPT better follow our instructions and make the most of the information provided in the prompt.

\textbf{Clear Content} To investigate the effect of our proposed clear prompt content, we conduct ablation study on (1) recall of tables and columns and (2) specify foreign keys. The results are recorded in Table~\ref{tab: ablation}. From this table, we can see that the performance of C3 decreases by 2.3\% when removing the operation of recalling of tables and columns; besides, the performance of C3 decreases by 2.6\%  when removing the operation of specifying foreign keys. It demonstrates that the operations of recalling tables and columns and specifying foreign keys, which provides clear prompt content, are necessary for C3. We further use Area Under ROC Curve (AUC) to evaluate the performance of table and column recall. As illustrated in Section~\ref{cp}, we only keep top-4 tables in the database and top-5 columns for each remained table to form a ranked schema sequence. The results are recorded in Table~\ref{tab: AUC}, which shows that the AUC of table recalling and column recalling are 0.9588 and 0.9833 respectively.

\subsection{Effect of Calibration of Model Bias}
To explore the effect of our proposed calibration method, Calibration with Hints (CH), we perform an ablation study on it. The experimental results are illustrated in Table~\ref{tab: ablation}. From this table, we can see that the execution accuracy decreases by 1.5\% when calibration is removed. It verifies that calibration of model bias is necessary for chatGPT-based Text-to-SQL.

\subsection{Effect of Self-Consistency}

\begin{figure}[t]
  \centering
  \includegraphics[width=1.0\linewidth]{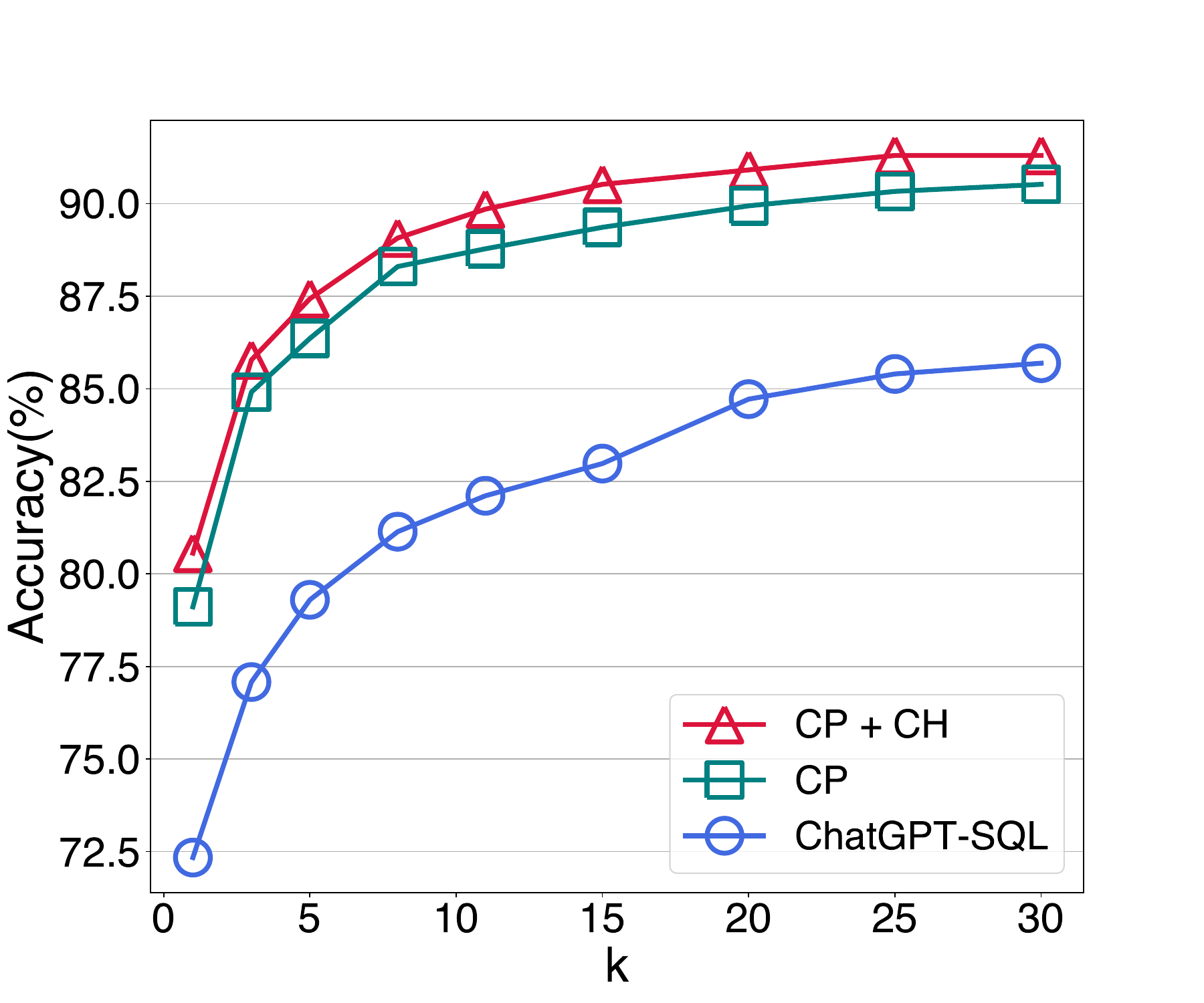}
  \caption{The accuracy of \textit{k} experiments under different prompts. For each sample in the dev set, if it gets the correct answer at least once in \textit{k} experiments, it is counted as correct. We gradually increase \textit{k} until it reaches 30.}
  \label{fig:coverage}
\end{figure}

\begin{figure*}
\centering
\begin{minipage}[t]{0.48\textwidth}
\centering
\includegraphics[width=1.0\textwidth]{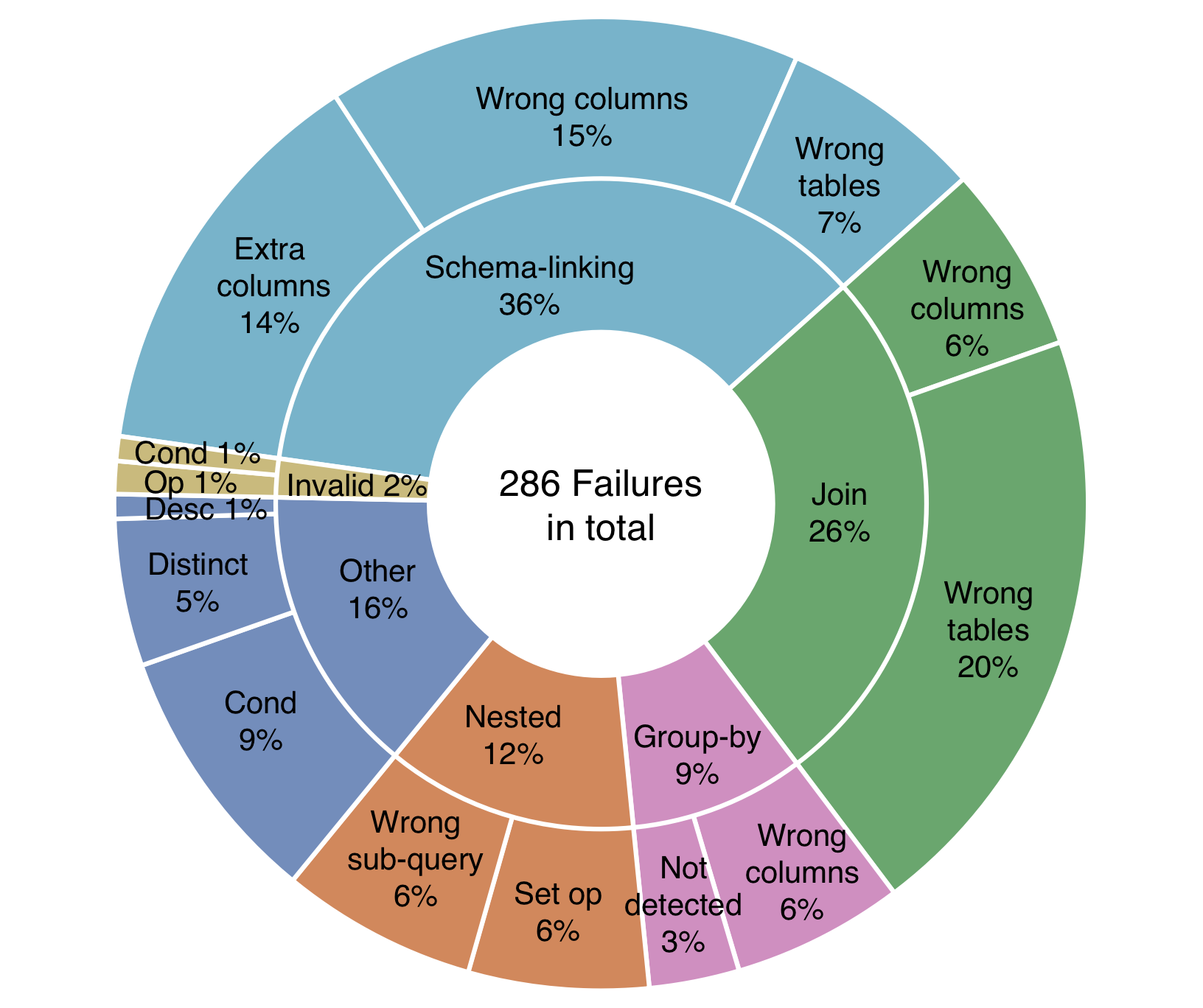}
\label{fig:failure_chatgpt-sql}
\hspace{0.1cm}
\subcaption{ChatGPT-SQL + ChatGPT}
\end{minipage}
\begin{minipage}[t]{0.48\textwidth}
\centering
\includegraphics[width=1.0\textwidth]{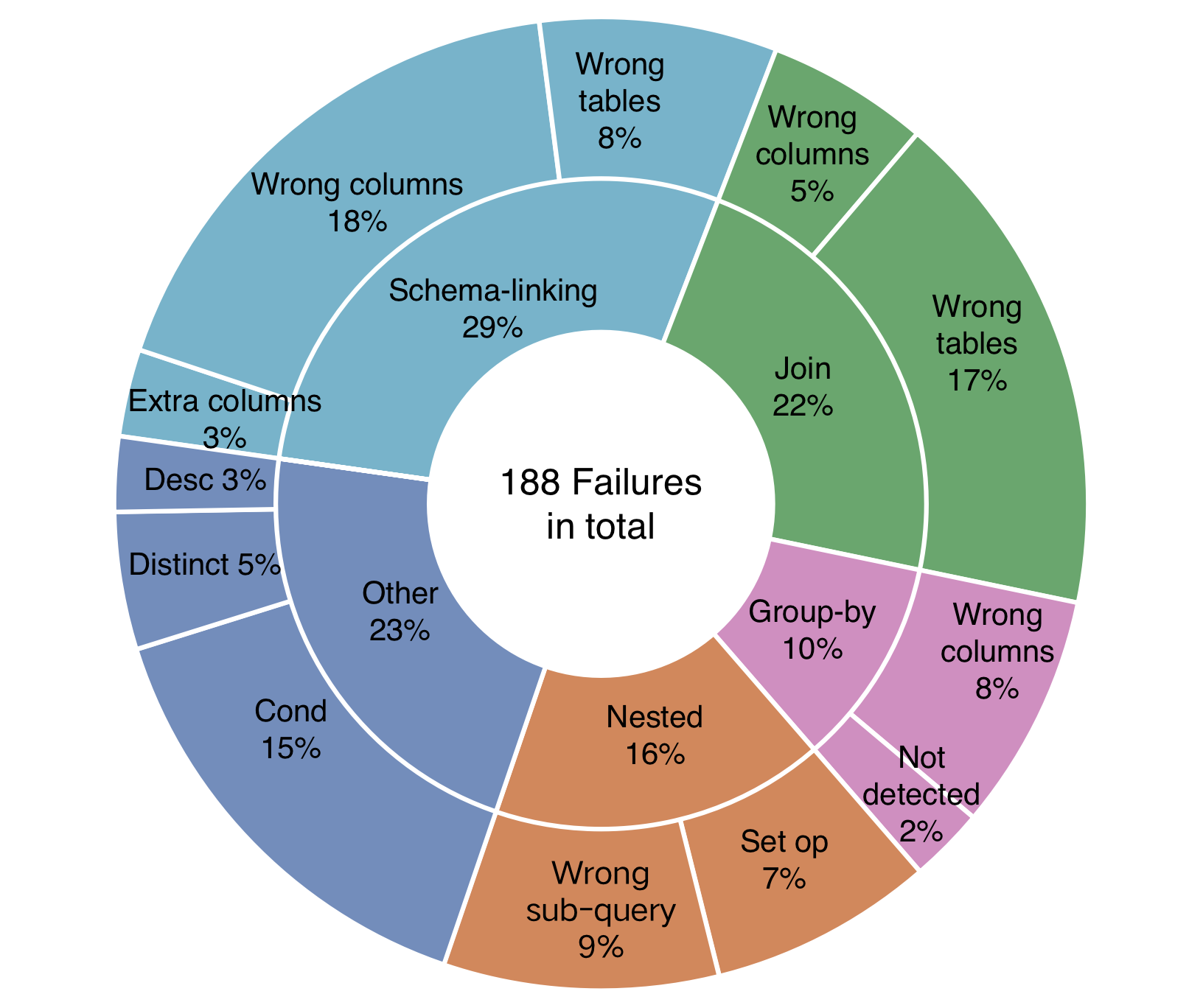}
\label{fig:failure_c3}
\subcaption{C3 + ChatGPT (Ours)}
\end{minipage}
\caption{Error Distributions of ChatGPT-SQL and C3 on dev set. There are a total of 286 failures in ChatGPT-SQL, and 188 failures in C3. If a sample contains errors of multiple categories, it will be counted in all these categories.}  \label{fig:failure_distribution}
\end{figure*}


In this section, we discuss about the effect of self-consistency. We first conduct an experiment to show ChatGPT's high potential on Text-to-SQL task. We perform $k$ independent experiments and record the changes of accuracy. The results are shown in Figure~\ref{fig:coverage}. For each question, if ChatGPT generates one or more correct SQL in the $k$ experiments, we think ChatGPT can generate a accurate SQL for the question. It can be observed that when using ChatGPT-SQL, the accuracy can reach approximately 85.7\% when $k=30$. However, the accuracy is only 72.3\% when $k=1$, resulting in a 13.4\% gap. This gap shows the enormous potential of ChatGPT. When using our two prompts, though the gaps are relatively smaller and the curves are flatter, the gap is still noticeable. However, We can find that a slight increase of $k$ can lead to a large increase of accuracy. This is due to the unstability of ChatGPT, which may not generate the correct answer every time. Therefore, as illustrated in~\ref{CO}, using self-consistency can utilize multiple answers to get the most consistent one and mitigate this unstability.  
In the experiment, we set the parameter $n$ to 20 in GPT-3.5 API call function in order to generate twenty SQL queries in a single run. Among these SQL queries, we cluster them based on their execution results and select one SQL query from the largest cluster as the final SQL. The ablation study of self-consistency is recorded in Table~\ref{tab: ablation}, we can see that self-consistency improves performance by 1.3\% and is therefore necessary to maintain the stable performance.

\subsection{Error Analysis}
To clearly demonstrate the effectiveness of our proposed method, we conduct an error analysis of ChatGPT-SQL and our C3 method on the dev set. We adopt the error classification method proposed in literature~\cite{pourreza2023din}. Specifically, we manually examined the errors occurred in the generated SQL answers and classified them into six categories, as shown in Figure~\ref{fig:failure_distribution}. 

Firstly, we can see that, by using our C3 method, the total number of errors decreases by approximately 34\% compared to ChatGPT-SQL. The most noteworthy reduction happens in the \textit{Schema-linking} category, which contains three subcategories. Among these subcategories, the subcategory \textit{Extra-columns} is mostly caused by Bias 1(refer to Section~\ref{CH}). We can see that the errors belonging to this subcategory has a proportional decrease of 11\%. It demonstrates the significant effectiveness of CH in calibrating the model bias.

Secondly, the subcategory \textit{Set op} in category \textit{Nested} demonstrates the misuse of operations like \textit{INTERSECT} and \textit{EXCEPT}, and the subcategory \textit{Distinct} in category \textit{Other} indicates repetitive execution results. Errors belonging to these subcategories are mostly caused by Bias 2 (refer to Section~\ref{CH}). Though the proportions of these categories change a little, the actual number of errors in these categories decreases. This indicates that CH can effectively calibrate Bias 2.

And finally, in the category \textit{Schema-linking} and \textit{Join}, we can see that the errors of choosing wrong tables and columns significantly decrease, which can be credited to the operation of table and column recall as well as specifying foreign keys in the proposed CP. As for the other categories, the number of errors is also reduced, which can be credited to the CO method for it improves the stability and maintains high performance in each experiment.
The overall reduction of errors is attributed to the combined effect of the C3 method.

\section{Conclusion}
This paper proposes a novel zero-shot Text-to-SQL method based on ChatGPT, called C3, which achieves the state-of-the-art zero-shot Text-to-SQL performance. C3 provides a systematic treatment for GPT-based Text-to-SQL from the perspective of model input, model bias, and model output. It is potential to forge a new trend in the GPT-based Text-to-SQL research.

\section{Acknowledgments}
We thank Yu Mi and Yilin Li for their valuable assistance with the experiments.


\bibliography{custom}
\bibliographystyle{acl_natbib}

\newpage

\onecolumn

\appendix
\section{Clear Prompting}
\label{sec:conversation}
Large language models (LLMs) are highly sensitive to the input prompt. A clear prompt can have a substantial positive impact on the performance of downstream tasks. This claim is empirically evidenced by our C3 prompt and corresponding experimental results in the Text-to-SQL task. Furthermore, we present the following question to ChatGPT for suggestions: \\

\label{sec:q1}
    \fbox{
    \parbox{1.0\textwidth}{
    \vspace{0.3em}
    \begin{restatable*}{*theorem}{Q1}
    \label{fig:Q1}
    \textbf{Human}: \textbf{Does using a clear prompt layout benefit ChatGPT in performing Text-to-SQL tasks?} \\
    
    \textbf{ChatGPT}: Yes, using a structured prompt style can greatly benefit ChatGPT in performing Text-to-SQL tasks. Text-to-SQL tasks involve converting natural language queries into SQL queries that can be executed on a database. These tasks often require a clear understanding of the query intent and the underlying database schema. \\

    Structured prompts provide a specific format or template that guides the model to generate SQL queries in a structured and predictable manner. By using a structured prompt style, you can specify the desired structure and components of the SQL query, such as the SELECT, WHERE, and JOIN clauses, as well as the table and column names. This helps to guide the model's generation process and improve the accuracy and consistency of the generated SQL queries. \\
    
    Structured prompts can also help overcome some of the challenges of language models, such as ambiguity and verbosity. By providing explicit instructions and constraints in the prompt, you can guide the model to generate more precise and concise SQL queries. This reduces the chance of the model producing queries that are syntactically correct but semantically incorrect or incomplete. \\
    
    However, it's important to note that while structured prompts can improve the performance of ChatGPT in Text-to-SQL tasks, they are not a complete solution. These tasks can be complex, and the model may still require additional training, fine-tuning, or post-processing techniques to achieve optimal performance. Additionally, structured prompts may limit the model's flexibility in handling novel or out-of-domain queries, so a balance between structure and flexibility should be considered when designing prompts for Text-to-SQL tasks.
    \end{restatable*}
    \vspace{0.3em}
    }
}
\\


\section{Schema Linking}
\label{sec:ranking}
\subsection{The example of table recall prompt} 
\label{sec:table_recall}
Given the database schema and question, perform the following actions: \\
1 - Rank all the tables based on the possibility of being used in the SQL according to the question from the most relevant to the least relevant, Table or its column that matches more with the question words is highly relevant and must be placed ahead.\\
2 - Check whether you consider all the tables.\\
3 - Output a list object in the order of step 2, Your output should contain all the tables. The format should be like: \\
\text{{[}}\\
"table\_1", "table\_2", ...\\
\text{{]}}\\

\noindent Schema:\\
\# continents ( contid, continent ) \\
\# countries ( countryid, countryname, continent ) \\
\# car\_makers ( id, maker, fullname, country ) \\
\# model\_list ( moddeli, maker, model ) \\
\# car\_names ( makeid, model, make ) \\
\# cars\_data ( id, mpg, cylinders, edispl, horsepower, weight, accelerate, year ) \\

\noindent Question: \\
\#\#\# What is the name of the different car makers who produced a car in 1970?\\

\subsection{The example of column recall prompt} 
\label{sec:column_recall}
Given the database tables and question, perform the following actions: \\
1 - Rank the columns in each table based on the possibility of being used in the SQL, Column that matches more with the question words or the foreign key is highly relevant and must be placed ahead. You should output them in the order of the most relevant to the least relevant.\\
Explain why you choose each column.\\
2 - Output a JSON object that contains all the columns in each table according to your explanation. The format should be like: \\
\{\\
    "table\_1": ["column\_1", "column\_2", ......], \\
    "table\_2": ["column\_1", "column\_2", ......],\\
    "table\_3": ["column\_1", "column\_2", ......],\\
     ......\\
\}\\

\noindent Schema:\\
\# car\_makers ( id, maker, fullname, country ) \\
\# model\_list ( modelid, maker, model ) \\
\# car\_names ( makeid, model, make ) \\
\# cars\_data ( id, mpg, cylinders, edispl, horsepower, weight, accelerate, year ) \\
Foreign keys: \\
\# model\_list.maker = car\_makers.id\\
\# car\_names.model = model\_list.model\\
\# cars\_data.id = car\_names.makeid\\

\noindent Question: \\
\#\#\# What is the name of the different car makers who produced a car in 1970?\\

\section{The examples of prompts}
\label{sec:example}

\subsection{Clear Layout} 

\#\#\# Complete sqlite SQL query only and with no explanation\\
\#\#\# Sqlite SQL tables, with their properties: \\
\# \\
\# stadium ( stadium\_id, location, name, capacity, highest, lowest, average ); \\
\# singer ( singer\_id, name, country, song\_name, song\_release\_year, age, is\_male ); \\
\# concert ( concert\_id, concert\_name, theme, stadium\_id, year ); \\
\# singer\_in\_concert ( concert\_id, singer\_id ). \\
\# \\
\#\#\# How many singers do we have? \\
SELECT

\subsection{Complicated Layout}
Complete sqlite SQL query only and with no explanation. 
How many singers do we have? Sqlite SQL tables, with their properties: singer : singer.singer\_id , singer.name , singer.country , singer.age , singer.song\_name , singer.song\_release\_year , singer.is\_male | stadium : stadium.location , stadium.name , stadium.capacity , stadium.highest , stadium.lowest , stadium.average , stadium.stadium\_id | concert : concert.theme , concert.year , concert.concert\_id , concert.concert\_name , concert.stadium\_id | singer\_in\_concert : singer\_in\_concert.concert\_id , singer\_in\_concert.singer\_id\\
SELECT

\subsection{C3 Prompt}
\label{sec:c3_prompt}
\noindent \textbf{Role:} \textcolor{teal}{SYSTEM} \\
\textbf{Content:} \\
You are now an excellent SQL writer, first I'll give you some tips and examples, and I need you to remember the tips, and do not make same mistakes.\\

\noindent \textbf{Role:} \textcolor{red}{USER} \\
\textbf{Content:} \\
Tips 1: \\
Question: Which A has most number of B?\\
Gold SQL: select A from B group by A order by count ( * ) desc limit 1;\\
Notice that the Gold SQL doesn't select COUNT(*) because the question only wants to know the A and the number should be only used in ORDER BY clause, there are many questions asks in this way, and I need you to remember this in the the following questions.\\

\noindent \textbf{Role:} \textcolor{blue}{ASSISTANT} \\
\textbf{Content:} \\
Thank you for the tip! I'll keep in mind that when the question only asks for a certain field, I should not include the COUNT(*) in the SELECT statement, but instead use it in the ORDER BY clause to sort the results based on the count of that field.\\

\noindent \textbf{Role:} \textcolor{red}{USER} \\
\textbf{Content:} \\
Tips 2: \\
Don't use "IN", "OR", "LEFT JOIN" as it might cause extra results, use "INTERSECT" or "EXCEPT" instead, and remember to use "DISTINCT" or "LIMIT" when necessary.\\
For example, \\
Question: Who are the A who have been nominated for both B award and C award?\\
Gold SQL should be: select A from X where award = 'B' intersect select A from X where award = 'C';\\

\noindent \textbf{Role:} \textcolor{blue}{ASSISTANT} \\
\textbf{Content:} \\
Thank you for the tip! I'll remember to use "INTERSECT" or "EXCEPT" instead of "IN", "NOT IN", or "LEFT JOIN" when I want to find records that match or don't match across two tables. Additionally, I'll make sure to use "DISTINCT" or "LIMIT" when necessary to avoid repetitive results or limit the number of results returned.\\

\noindent \textbf{Role:} \textcolor{red}{USER} \\
\textbf{Content:} \\
\#\#\# Complete sqlite SQL query only and with no explanation, and do not select extra columns that are not explicitly requested in the query. \\
 \#\#\# Sqlite SQL tables, with their properties: \\
\# \\
\# singer ( singer\_id, name, country, age )\\
\# stadium ( capacity, highest, lowest, average )\\
\# concert ( theme, year, concert\_id, concert\_name )\\
\# singer\_in\_concert ( concert\_id, singer\_id )\\
\# concert.stadium\_id = stadium.stadium\_id\\
\# singer\_in\_concert.singer\_id = singer.singer\_id\\
\# singer\_in\_concert.concert\_id = concert.concert\_id\\
\# \\
\#\#\# How many singers do we have?\\
SELECT

\end{document}